\documentclass[letterpaper, 10 pt, journal, twoside]{IEEEtran}
\usepackage{cite}
\usepackage{bm}
\usepackage{amsmath,amssymb,amsfonts}
\usepackage{textcomp}
\usepackage{rotating}
\usepackage[export]{adjustbox}
\usepackage{multicol}
\usepackage{booktabs}
\usepackage{multirow}
\usepackage{xcolor}
\usepackage{colortbl}
\usepackage{pgf}
\usepackage{tabularx}
\usepackage{arydshln}
\usepackage{algorithm}
\usepackage{algpseudocode}
\usepackage{caption}
\usepackage{stfloats}
\usepackage{pdfpages}
\usepackage{graphicx}
\usepackage{siunitx}
\usepackage{nicematrix}
\usepackage{ctable}

\usepackage{xifthen}
\usepackage{lipsum} 

\newcounter{reviewer}
\newcounter{point}[reviewer]
\renewcommand{\thepoint}{\thereviewer.\arabic{point}} 

\newcommand{\shortreply}[2][]{\medskip \noindent \begin{sf}\textbf{Reply}:\  #2
		\ifthenelse{\equal{#1}{}}{}{ \hfill \footnotesize (#1)}%
		\medskip \end{sf}}

\usepackage{epsfig} 
\usepackage{mathptmx} 
\usepackage{times} 
\usepackage{amsmath} 
\usepackage{amssymb}  
\usepackage{varioref}
\usepackage[hidelinks]{hyperref}
\usepackage[capitalise,nameinlink,noabbrev]{cleveref}
\newlength{\Oldarrayrulewidth}

\definecolor{1}{RGB}{0, 255, 0}
\definecolor{2}{RGB}{94, 243, 0}
\definecolor{3}{RGB}{129, 230, 0}
\definecolor{4}{RGB}{155, 217, 0}
\definecolor{5}{RGB}{176, 204, 0}
\definecolor{6}{RGB}{193, 189, 0}
\definecolor{7}{RGB}{208, 174, 0}
\definecolor{8}{RGB}{221, 158, 0}
\definecolor{9}{RGB}{231, 142, 0}
\definecolor{10}{RGB}{240, 124, 0}
\definecolor{11}{RGB}{247, 104, 0}
\definecolor{12}{RGB}{252, 82, 0}
\definecolor{13}{RGB}{254, 55, 0}
\definecolor{14}{RGB}{255, 0, 0}

\definecolor{1}{RGB}{240, 249, 33}
\definecolor{2}{RGB}{250, 216, 36}
\definecolor{3}{RGB}{254, 184, 44}
\definecolor{4}{RGB}{250, 155, 61}
\definecolor{5}{RGB}{241, 129, 77}
\definecolor{6}{RGB}{228, 105, 94}
\definecolor{7}{RGB}{212, 82, 112}
\definecolor{8}{RGB}{194, 60, 129}
\definecolor{9}{RGB}{172, 38, 148}
\definecolor{10}{RGB}{146, 15, 163}
\definecolor{11}{RGB}{119, 1, 168}
\definecolor{12}{RGB}{88, 1, 164}
\definecolor{13}{RGB}{55, 4, 153}
\definecolor{14}{RGB}{13, 8, 135}

\definecolor{1}{RGB}{240, 249, 33}
\definecolor{2}{RGB}{248, 225, 37}
\definecolor{3}{RGB}{253, 202, 38}
\definecolor{4}{RGB}{253, 180, 47}
\definecolor{5}{RGB}{251, 159, 58}
\definecolor{6}{RGB}{245, 139, 71}
\definecolor{7}{RGB}{237, 121, 83}
\definecolor{8}{RGB}{227, 104, 95}
\definecolor{9}{RGB}{216, 87, 107}
\definecolor{10}{RGB}{204, 71, 120}
\definecolor{11}{RGB}{189, 55, 134}
\definecolor{12}{RGB}{173, 39, 147}
\definecolor{13}{RGB}{156, 23, 158}
\definecolor{14}{RGB}{136, 8, 166}

\usepackage[T1]{fontenc}
\ifCLASSINFOpdf
\else
\fi
\begin{document}
		%
		\title{Hybrid Learning of Time-Series Inverse Dynamics Models for Locally Isotropic Robot Motion}
		%
		%
		%
		
		\author{Tolga-Can~\c{C}allar$^1$ and
			Sven~B{\"o}ttger$^1$
		
			\thanks{Manuscript received: July, 29, 2022; Revised October, 12, 2022; Accepted November, 4, 2022.}
			\thanks{This paper was recommended for publication by Editor Cl{\'e}ment Gosselin upon evaluation of the Associate Editor and Reviewers' comments.
			}
			\thanks{$^1$Tolga-Can~\c{C}allar and
			Sven~B{\"o}ttger are with the Institute for Robotics and Cognitive Systems, Universit{\"a}t zu L{\"u}beck, Germany {\tt\footnotesize callar@rob.uni-luebeck.de}}

			\thanks{Digital Object Identifier (DOI): 10.1109/LRA.2022.3222951}
		}
		
		%
		%

	\markboth{IEEE Robotics and Automation Letters. Preprint Version. Accepted November, 2022}
	{\c{C}allar \MakeLowercase{\textit{et al.}}: Hybrid Learning of Time-Series Inverse Dynamics Models for Locally Isotropic Robot Motion} 
	%



	\maketitle
	
	\begin{abstract}
		Applications of force control and motion planning
		often rely on an inverse dynamics model to represent the high-dimensional dynamic behavior of robots during motion. The widespread
		occurrence of low-velocity, small-scale, locally isotropic motion (LIMO) typically
		complicates the identification of appropriate models due to
		the exaggeration of dynamic effects and sensory
		perturbation caused by complex friction and phenomena of hysteresis, e.g., pertaining to joint
		elasticity. We propose a hybrid
		model learning base architecture combining a rigid
		body dynamics model identified by parametric regression and
		time-series neural network architectures based on multilayer-perceptron, LSTM, and Transformer topologies.
		Further, we introduce a novel joint-wise rotational history
		encoding, reinforcing temporal information to effectively model
		dynamic hysteresis. The models are evaluated on
		a KUKA iiwa 14 during algorithmically generated locally
		isotropic movements. Together with the rotational encoding, the proposed architectures 
		outperform state-of-the-art baselines by a magnitude of 10$\bm^3$ yielding an RMSE of 0.14 Nm. Leveraging the hybrid structure
		and time-series encoding capabilities, our approach allows for
		accurate torque estimation, indicating its applicability 
		in critically force-sensitive applications during motion sequences exceeding the capacity of conventional inverse dynamics models while retaining trainability in face of scarce data and explainability due to the employed physics model prior.
	\end{abstract}

	\begin{IEEEkeywords}
		Dynamics, Model Learning for Control, Deep Learning Methods, Force and Tactile Sensing, Force Control
	\end{IEEEkeywords}

	\IEEEpeerreviewmaketitle

	\section{Introduction}

	\IEEEPARstart{T}{he} availability of an inverse dynamics model (\textbf{IDM}) that infers the mechanical forces within a robotic system from its motion constitutes the foundation for a multitude of applications within the area of force control and motion planning. With the advent of multi-articulated robots employed in real-world tasks under uncontrolled conditions, there is an increased necessity for modeling strategies that are able to accurately and robustly represent the high dimensional dynamics of mechanically complex robots executing a kinematically and dynamically wide range of motions.
 
	In real-world manipulation and physical interaction applications, the executed robot motion is often comprised of segments of low velocity and local isotropy in the form of frequent reversals of the joint-wise directions of rotation within small angular intervals for which we introduce the umbrella term of \textbf{Locally Isotropic Motion (LIMO)}.
	\begin{figure}[t]
		\centering
		\includegraphics[width=1\columnwidth]{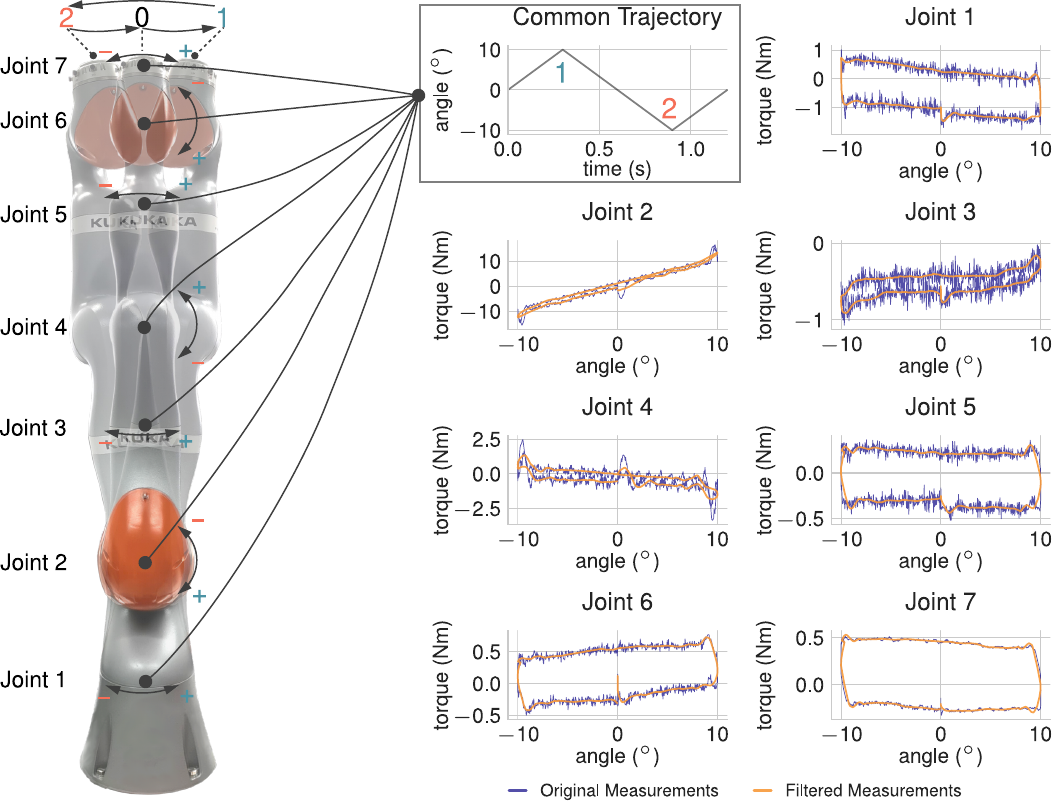}
		\caption{Joint torque hysteresis on the KUKA LBR iiwa 14 during locally isotropic motion (LIMO) in the form of a joint-wise common cyclical linear trajectory. Also, note the presence of backlash torques at directional reversals.}
		\label{hysteresis}
	\end{figure}
	LIMO can be encountered in collaborative, medical, social, or field robotics, where movements may predominantly manifest as irregular fine-scale motion sequences for precise compliant manipulation, e.g. in robotic sonography \cite{callar}, and reactive contact control during continuous physical interactions with an uncontrolled environment and human agents \cite{8206543}.
	This is distinct from locally anisotropic motion, e.g. in pick-and-place tasks \cite{hitzler} or large amplitude sinusoidal motion \cite{Sturz}, where joint-wise rotation directions remain constant for larger angular intervals, i.e. saturating deformation states of flexible internal components and interfaces until, effectively, mechanical rigidity is approached up to a further directional reversal.
	During LIMO, an increased dynamic influence of hysteretic effects pertaining to low-velocity friction \cite{bona2005friction}, joint flexibility, and backlash can be observed \cite{8206543,allgeier}. As shown by the formation of characteristic joint torque hysteresis loops in Fig. \ref{hysteresis}, the sensed torques substantially depend on the rotation history, which is apparent from non-linearly increasing and saturating torques during rotation, and torque deflections at rotational reversals, which is consistent with mechanical hysteresis.
	
	As such, LIMO quickly leads to the summation of multiple hysteretic torque deviations. In the context of IDMs, this introduces a non-negligible torque contribution that, albeit deterministic, results from the system's dynamic state history and is therefore difficult to model by conventional, steady-state approaches \cite{rueckert}. Together with inherently complex non-linear dynamics of common robots, this aggravates the IDM problem for the deployment of robotics in critically force-sensitive applications where this motion type occurs.

	IDM techniques are classically composed as parametric models \cite{Sturz} derived from the seminal work presented in \cite{atkeson}, which formulates the Newton-Euler equations of motion linearly with respect to the inertial link parameters to identify a model by linear regression from dynamic measurements.
	
	This so-called rigid body dynamics (\textbf{RBD}) model is often extended to also include basic frictional parameters \cite{KHALIL2002191}. 

	For collaborative applications, lumped dynamic models in the form of momentum observers are proposed, disregarding dynamics related to acceleration trading faster model predictions and independence from noisy acceleration measurements for lower prediction accuracy \cite{haddadin}.
	Although providing good model explainability and robustness, parametric model formulations are inherently incomplete in face of a myriad of dynamic effects exceeding closed-form model definitions, and thus often unable to be encode temporal information and yield accurate estimates during challenging motion such as LIMO.
	On the other end of the spectrum, black-box techniques are investigated, ranging from non-parametric, Gaussian processes-based methods to artificial neural networks \cite{nguyen,vijayakumar2000locally,yilmaz,polyodros} employed to learn models directly from data, which in theory allows for the approximation of any non-linear dynamic system, although conversely introducing the risk of model variance and unexplainability. 
	Thus, grey-box model learning approaches have been proposed, e.g, in form of neural network and Gaussian Process regression incorporating formulations of Newtonian, Lagrangian and Hamiltonian mechanics, where physical prior information is either directly embedded in a data-driven regression method, i.e. as a differentiable computational graph \cite{ledezma,lutter2019deep,greydanus2019hamiltonian,nguyen} or used separately in an error-learning scheme \cite{hitzler,nguyen,Calandraetal2015}.
	However, time-dependency effects, innate to robot dynamics, have often been neglected. Advancements in this respect were made by casting the dynamic modeling problem as a time-series regression task and using recurrent neural networks for IDMs\cite{rueckert}.

	The commonality between these previous approaches is nonetheless that the investigated motion types are mainly intended for and validated on locally anisotropic motion, as defined above. 
	To explore methods appropriate for the modeling of the dynamics of LIMO, we design dedicated techniques for LIMO modeling and present the following contributions:
	\begin{itemize}
		\item A hybrid learning base architecture combining a rigid body dynamics model prior learned by parametric identification with data-driven model learning using neural network topologies based on multilayer perceptrons and time-series networks in form of LSTM and Transformer topologies. (Sec. \ref{hybridModelLeraningBaseArchitecture})
		\item A novel joint rotation history encoding as a model input feature to reflect torque hysteresis states. (Sec. \ref{encoding})
		\item A motion generation algorithm to obtain dynamically versatile training data for LIMO. (Sec. \ref{trainingMotion})
		\item A comparative evaluation of various state-of-the-art IDM techniques on long-term proprioceptive time-series measurements acquired on the KUKA LBR iiwa 14 during LIMO and conventional motion. (Sec. \ref{eval})
		\item Our model implementations and LIMO data sets are shared with the research community in \cite{repo}.
	\end{itemize}
	
	The rationale behind our proposed methodology is based on two assumptions: Firstly, as evident from the mentioned related work on grey-box modeling the hybridization of a semantically pre-defined parametric model with a universal approximator should create synergetic effects regarding explainability, prediction robustness, and representation capabilities to eventually improve the modeling of complex dynamics occurring during challenging motion types as LIMO. Secondly, leveraging sequence history information on multiple levels, e.g. using different time-series neural network architectures and sequence encodings should benefit model accuracy, especially when dynamic states similar to LIMO are targeted.

	Honing in on the outlined dynamic modeling problem, principal questions addressed by our work revolve around the infamous bias-variance tradeoff to design the optimal model architecture:
	\textbf{1)} Can rigid body dynamics models sufficiently represent the dynamics of LIMO? \textbf{2)} Can neural networks provide higher accuracy while generalizing to out-of-distribution dynamic states, given resource-intensive data acquisition in robotics? \textbf{3)} What is the benefit provided by a hybrid approach? \textbf{4)} How important is temporal information for model accuracy? 
	
	After delineating the fundamentals of the IDM problem, we present the details of our model architectures, feature design, data generation, and evaluation process.

	\section{Robot Dynamics}
	Considering a serial kinematic structure composed of ${n=7}$ joints, in the case of our reference platform the KUKA iiwa 14, we define the dynamics by a composite model  
	\begin{align}
		\bm{\tau} = \overbrace{ \underbrace {H(\bm{q})\bm{\ddot{q}}+C(\bm{q},\bm{\dot{q}})\bm{q}+g(\bm{q})}_{\bm{\tau}_\mathrm{RBD}}+
			\underbrace{\epsilon(\bm{q},\bm{\dot{q}}, \bm{\ddot{q}},\star)}_{\bm{\tau}_\mathrm{ADD}}\label{modelFunction}}^{D(\bm{q},\bm{\dot{q}},\bm{\ddot{q}},\star)}\;,
	\end{align}
	expressing the joint torques $\bm{\tau}\in \mathbb{R}^{{n=7}}$ as a function of the kinematic state variables of joint positions and their temporal derivatives, $\bm{q},\bm{\dot{q}}, \bm{\ddot{q}}\in\mathbb{R}^{{n}}$, along with an undefined set of additional variables subsumed under $\star$. With $H(\bm{q})\in\mathbb{R}^{n\times n}$ as the symmetric and positive-definite joint-space inertia matrix, $C(\dot{\bm{q}},\ddot{\bm{q}})\in\mathbb{R}^{n\times n}$ as the combination of the Coriolis and centripetal forces and $g(\bm{q})$ as the gravity vector, the first terms grouped under $\bm{\tau}_{RBD}$ represent the torques solely governed by the framework of rigid body dynamics \cite{featherstone1987inverse}. 
	Due to the presence of a plethora of effects that exceed those simplifying assumptions, e.g. manifestations of complex friction dynamics, e.g. \textit{Stribeck} and \textit{LuGre} friction \cite{bona2005friction}, and joint flexibility, the model also includes a collective compensatory term $\epsilon(\bm{q},\bm{\dot{q}}, \bm{\ddot{q}},\star)$ to accommodate for joint torques $\bm{\tau}_{ADD}$ resulting from the entirety of additional phenomena affecting the true robot dynamics. This also includes perturbations in the technical perception of the model variables caused by arbitrary changes of the actual or sensed dynamic state, e.g., deviations in the measurement of joint positions and computationally derived velocities and accelerations, or the sensed joint torques themselves, i.e. mainly due to joint flexibility, friction or vibration \cite{8206543}.
	
	We define the IDM problem as the identification of a dynamic model function ${F(\bm{q},\bm{\dot{q}},\bm{\ddot{q}},\bm\star)}$ that approximates the observable true dynamics $D$ such that the model estimates $\bm{\tau}_\mathrm{est}$ fulfill
	\begin{align}
		\lvert\lvert \underbrace{D(\bm{q},\bm{\dot{q}},\bm{\ddot{q}},\bm\star)}_{\bm\tau} - \underbrace{F(\bm{q},\bm{\dot{q}},\bm{\ddot{q}},\bm\star)}_{\bm{\tau}_\mathrm{est}}
		\rvert\rvert \rightarrow 0
	\end{align}
	over a maximum range of the dynamic state space, where,  in practical terms, $\bm{q}$ and $\bm{\tau}$ are observed via position sensors and strain-gauge sensors located at the output side of each joint of the KUKA LBR iiwa 14.
	\section{Hybrid Model Learning Base Architecture}\label{hybridModelLeraningBaseArchitecture}
\begin{figure}[b]
		\centering
		\includegraphics[width=\columnwidth]{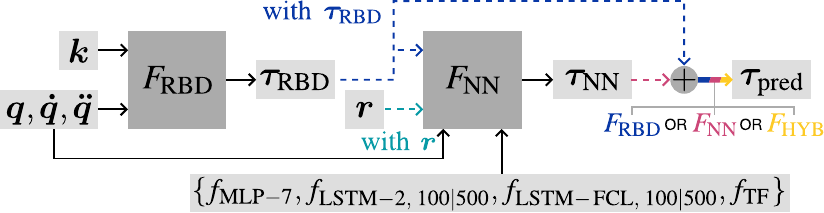}
		\caption{Directed Graph of the Hybrid Base Architecture within which the dashed connections are ablated and the neural network configurations $F_{\mathrm{NN}}$ are varied as part of the evaluation experiments (see Sec. \ref{eval}).}\vspace{-.25cm}
		\label{fig:NetArch}
	\end{figure}

	Exploiting the compositionality of the model formulation in (\ref{modelFunction}), we propose a hybrid model base architecture $F_\mathrm{HYB}$ consisting of two sequential elements, beginning with a parametric rigid body dynamics model $F_\mathrm{RBD}$ and a downstream neural network-based model $F_\mathrm{NN}$ as per Fig. \ref{fig:NetArch} and:
	\begin{align}\label{hybridArchitecture}
		\underbrace{F_\mathrm{HYB}}_{\bm{\tau}_\mathrm{HYB}} = \underbrace{F_\mathrm{RBD}(\bm{q},\bm{\dot{q}},\bm{\ddot{q}},\bm{k})}_{\bm{\tau}_\mathrm{RBD}}+\underbrace{F_\mathrm{NN}({\bm{q},\bm{\dot{q}}},\bm{\ddot{q}},\bm{r},\bm\tau_\mathrm{RBD})}_{\bm{\tau}_\mathrm{NN}}\;.
	\end{align}
	
	$F_\mathrm{HYB}=\bm\tau_\mathrm{HYB}$ is the sum of the predictions made by both model components $F_\mathrm{RBD} = \bm\tau_\mathrm{RBD}$ and $F_\mathrm{NN} = \bm\tau_\mathrm{NN}$, where $F_\mathrm{NN}$ itself also depends on the prediction $F_\mathrm{RBD} = \bm\tau_\mathrm{RBD}$ as an auxiliary input. As such, $F_\mathrm{HYB}$ constitutes a hybrid between a physics-based model prior and a data-driven, learned neural network. The internal structure of the respective components along with the definition of the utilized input features is the subject of the following.
	\subsection{Upstream Parametric Model Identification}
	Initially, we apply a parametric identification procedure based on the Newtonian formulation of mechanics to obtain a model $F_\mathrm{RBD}$ by statistical regression of the inertial link parameters implicitly embedded in the first terms of (\ref{modelFunction}) pertaining to the torques $\bm{\tau}_\mathrm{RBD}$ that result from rigid body dynamics (RBD).
	For that, we extend the rotational component of the Newton-Euler equations to include viscous and Coulomb friction between adjacent links as a simplified actuation friction model \cite{KHALIL2002191}, and define the torque $\bm{\tau}_{ii}$ acting upon the $i$-th link due to its own motion as
	\begin{flalign}
		\begin{adjustbox}{width=.9\columnwidth}
			$\bm{\tau}_{ii} = (\bm{g} - \bm{\ddot{p}}_i) \times m_i\bm{c}_i +
			\bm{I}_i\bm{\dot{\omega}}_i+\bm{\omega}_i\times(\bm{I}_i\bm{\omega}_i)+F_i^v\bm{\dot{\omega}}_i+F_i^c\text{sign}(\bm{\dot{\omega}}_i)\;,$
		\end{adjustbox}
	\end{flalign}
	where $\bm{g}$ is the vector of gravitational acceleration, $\bm{\ddot{p}}_i$ is the linear link acceleration and $\bm{\omega}_i$ and $\bm{\dot{\omega}}_i$ are the angular link velocity and acceleration vectors. The parameters $m_i$, $\bm{c}_i$, $\bm{I}_i$, $F^v_i$ and $F^c_i$, constitute the link mass, center of mass, symmetric inertia tensor, viscous and Coulomb friction coefficients respectively.
	We reformulate the Newton-Euler equations to express the joint torques linearly in relation to the link parameters as presented in \cite{atkeson}, i.e.
	\begin{align}
		\label{torque}
		\bm{\tau}_{ii}=
		&\begin{bmatrix}
			\bm{0}&[(\bm{g}-\bm{\ddot{p}}_i)\times]&[\cdot\bm{\dot{\omega}}_i]+
			[\bm{\omega}_i
			\times][\cdot\bm{
				\dot{\omega}}_i]&\bm{\dot{\omega}}_i&\text{sign}(\bm{\dot{\omega}}_i)
		\end{bmatrix}\times\nonumber\\
		&{\underbrace{\setlength\arraycolsep{3pt}
				\begin{bmatrix}
					m_i&
					m_i\bm{c}_i&
					I_{xx}&
					I_{xy}&
					I_{xz}&
					I_{yy}&
					I_{yz}&
					I_{zz}&
					F_i^v&
					F_i^c
			\end{bmatrix}}_
			{\bm{\Phi}_i}}^\top\;.
\end{align}

\begin{figure}[t]
	\vspace{0mm}
	\centerline{\includegraphics[width=1.\columnwidth]{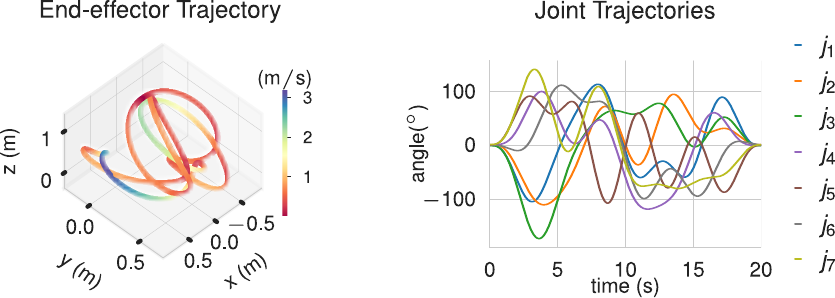}}
	\caption{(\emph{Left}) Cartesian path of the end-effector during the execution of the optimized excitation trajectory. (\emph{Right}) Excitation trajectory in configuration space.}
	\label{excitationTrajectory}
\end{figure}

With the Recursive Newton-Euler algorithm \cite{featherstone1987inverse}, given the kinematic structure $\bm{k}$, the link torques are transmitted backward along the kinematic chain to obtain a linear equation system
\begin{align}\label{linearEquation}
	\bm{\tau}_\mathrm{RBD}=\bm{K}(\bm{q}_,\bm{\dot{q}},\bm{\ddot{q}},\bm{k})\;\bm{\Phi}\;,
\end{align}
yielding $\bm{\tau}_\mathrm{RBD}$ for an observation matrix $\bm{K}$ constructed from a measured kinematic state via the link parameters $\bm\Phi \in \mathbb{R}^{12n\times1}$ \cite{atkeson}.
Based on this, we construct a stacked version of (\ref{linearEquation}) from $N$ observations pairs of the kinematic state $(\bm{q}_\mathrm{meas},\bm{\dot{q}}_\mathrm{meas},\bm{\ddot{q}}_\mathrm{meas})$ and the joint torque observations $\bm{\tau}_\mathrm{meas}$ to estimate $\bm\Phi$ by the ordinary least squares method from
\begin{align}\label{stackedSystem}
	\overline{\bm{\tau}}_\mathrm{meas} &= \overline{\bm{K}}(\bm{q}_\mathrm{meas},\bm{\dot{q}}_\mathrm{meas},\bm{\ddot{q}}_\mathrm{meas},\bm{k})\;\overline{\bm{\Phi}}\;,
\end{align}
where $\overline{\bm{\tau}}_\mathrm{meas} \in \mathbb{R}^{N\cdot n\times1}$, $\overline{\bm{K}}(\bm{q}_\mathrm{meas},\bm{\dot{q}}_\mathrm{meas},\bm{\ddot{q}}_\mathrm{meas},\bm{k}) \in \mathbb{R}^{N\cdot n\times12n}$ and $\overline{\bm{\Phi}} \in \mathbb{R}^{N\cdot 12n\times1}$.
To optimize identifiability of $\bm\Phi$, the kinematic observations are made while executing joint-wise disjunct sinusoidal dynamic excitation trajectories that minimize $\mathrm{cond}(\overline{\bm{K}}(\bm{q}_\mathrm{meas},\bm{\dot{q}}_\mathrm{meas},\bm{\ddot{q}}_\mathrm{meas},\bm{k}))$ conversely maximizing parametric identifiability \cite{Swevers,Sturz}. The obtained excitation trajectory (Fig. \ref{excitationTrajectory}) yields $\mathrm{cond}(\overline{\bm{K}}= 68)$.

Before constructing (\ref{stackedSystem}), the measurements are low-pass filtered to reduce a) noise amplification from numeric differentiation of joint position measurements to retrieve the velocities and accelerations and b) torque vibrations caused by actuation transmission \cite{8206543}. By rank reduction of $\overline{\bm{K}}(\bm{q}_\mathrm{meas},\bm{\dot{q}}_\mathrm{meas},\bm{\ddot{q}}_\mathrm{meas},\bm{k})$ to $\overline{\bm{K}}_b$ with $\mathrm{rank}(\overline{\bm{K}}_b)=57$ via QR-decomposition \cite{KHALIL2002191}, the reliably identifiable base parameters $\bm{\Phi}_b^\mathrm{LS}$ are determined and solved for by pseudo-inversion of the full-ranked version of the linear equation system (\ref{stackedSystem}). This yields the model
\begin{align}\label{rigidBodyModel}
	F_\mathrm{RBD} =  \bm{K}_b(\bm{q}_\mathrm{meas},\bm{\dot{q}}_\mathrm{meas},\bm{\ddot{q}}_\mathrm{meas},\bm{k})\bm{\Phi}_b^\mathrm{LS}=\bm{\tau}_\mathrm{RBD}\;,
\end{align}
which achieved a joint-wise average MSE of $0.2543 \text{ Nm}$ on the data recorded during the excitation experiment.

\subsection{Learning Downstream Models with Neural Networks}\label{baseArchitecture}
First, we introduce a motion encoding feature definition to alleviate the encoding of time-dependency information in the neural network input space. The subsequent sections \ref{MLP} to \ref{ioNorm} examine architectural variations of the neural network component $F_\mathrm{NN}$ utilized as the second stage of our hybrid model (\ref{hybridArchitecture}), for which we investigated and adapted several neural network types and topologies.

\subsubsection{\textbf{Rotational History Encoding}}\label{encoding}
Our work focuses on dynamic sequence-history dependencies, i.e., hysteretic behavior, that occur during LIMO. From a topological perspective, learning of such sequence models can be accomplished either through the exploitation of recurrence (Sec. \vref{LSTM}) or via an attention mechanism (Sec. \vref{Transformer}). Still, the sequence length over which time-dependencies can be effectively modeled is limited.
Conversely, generic feed-forward networks (Sec. \vref{MLP}) have no inherent means for sequence modeling.

Assuming that dynamic time-dependency effects mostly pertain to mechanical hysteresis during LIMO, we conjecture a correlation between observable joint torques and the history of joint positions that is defined by an associated hysteresis loop (Fig. \ref{hysteresis}). Hence, sufficient continuous rotation in one direction transitions the hysteresis into linear saturation,  and contrarily, frequent reversals of motion direction, i.e., during LIMO, confine the associated dynamic effects to manifest with pronounced non-linearity inside the hysteresis loop. To abstractly express the hysteretic state of the joint torques depending on the history of past rotations, we introduce a novel rotational history encoding for which we define a feature vector  
\begin{align}
	\bm{r} = \begin{bmatrix}r_1&r_2&\ldots&r_n \end{bmatrix},\; r_j \in [-10^\circ,10^\circ]
\end{align}
containing the signed total rotation angle of each joint since the respective last reversal of rotation direction. Before constructing $\bm{r}$ for a time-step, the joint angles buffered since the last reversal are squashed into the range $ \lvert\lvert\bm{r}^\circ\rvert\rvert < 10^\circ$ to exclude values redundantly encoding information for states where effects mechanical hysteresis is assumed to be saturated (Fig. \ref{hysteresis}). We hypothesize this auxiliary feature to aid sequence modeling networks with the extraction of long-term temporal dependencies within the data, as well as augment the performance of non-recurrent architectures by explicitly providing temporal information. The evaluation of the influence of this additional feature in the context of the generated data set for LIMO is presented in \ref{trainingMotion}.

$\bm{r}$ is concatenated with the remaining input features of the kinematic state $(\bm{q}_\mathrm{meas},\bm{\dot{q}}_\mathrm{meas},\bm{\ddot{q}}_\mathrm{meas})$ and the RBD model predictions $\bm\tau_\mathrm{RBD}$ (\ref{hybridArchitecture}), yielding the full input feature space 
\begin{align}\label{inputSpace}
	\bm{x}=\mathrm{ vec}(\bm{q},\bm{\dot{q}},\bm{\ddot{q}},\bm{r},\bm{\tau}_\mathrm{RBD}) \in \mathbb{R} ^{35}
\end{align}
for $F_\mathrm{NN}$ in (\ref{hybridArchitecture}) for which the network architecture variants are discussed next.
\subsubsection{\textbf{Multilayer Perceptron (MLP)}}\label{MLP}
Related to the topologies proposed by \cite{hitzler,yilmaz}, we construct a multilayer perceptron $ \mathrm{ MLP}_j $ for each robot joint $j$ consisting of
\begin{itemize}
	\item an input layer with 35 neurons accepting $ \bm{x} \in \mathbb{R} ^{35} $
	\item one hidden layer with 100 neurons with ReLu activation,
	\item an output layer with 1 neuron corresponding to the torque prediction $ f_{\mathrm{MLP}_j}={\tau}^j_\mathrm{NN} $ for the $j$-th joint.
\end{itemize}
The MLPs are arranged in a shared network topology such that the single-joint predictions are concatenated to $f_\mathrm{MLP}=\bm{\tau}_\mathrm{NN} = \begin{bmatrix}f_\mathrm{MLP_1}(\bm{x})&\ldots&f_\mathrm{MLP_{n=7}}(\bm{x}) \end{bmatrix} $ and the respective MLPs are trained jointly based on this output. For our implementation for the KUKA LBR iiwa with 7 joints, we abbreviate this architecture as \textbf{MLP-7}.
\subsubsection{\textbf{Long Short-Term Memory (LSTM)}}\label{LSTM}
Consider the presence of temporal dependency effects extending over comparatively long time horizons in the dynamic system, that cannot be sufficiently encoded by temporal derivation of positional measurements or even time encodings. Hence, the model is extended from a single time-step prediction to the mapping of a time-series of kinematic states to a single dynamic state prediction using long short-term memory networks inherently suited for sequential data \cite{rueckert}, i.e.,
\begin{align}
	f_\mathrm{LSTM}(\bm{x}({t_1}),\bm{x}({t_2}),\ldots,\bm{x}({T})) = \bm{\tau}_\mathrm{NN}(T)\;.
\end{align}
Here, $ T $ is the length of a sequence of observations $ \bm{x}(t) = \mathrm{vec}\left(\bm{\tau}_\mathrm{RBD}(t),\bm{q}(t),\bm{\dot{q}}({t}),\bm{\ddot{q}}(t) ,\allowbreak\bm{r}({t})\right)$ for every time-step $t_1,\ldots, T$, and $\bm{\tau}_\mathrm{NN}(T)\in\mathbb{R}^7$ is the joint torque prediction made at the last time-step $t=T$ based on the preceding kinematic observation sequence. We compare two sequence lengths $ T=100 $ and $ T=500 $, to elucidate whether longer time-frames improve prediction accuracy, or put differently, whether dynamic effects with longer temporal dependency significantly influenced the model. The network is implemented as two consecutive LSTM cells with forget gates where
\begin{itemize}
	\item the first LSTM cell has 35 input neurons accepting $ \bm{x}({t}) = \mathrm{vec}(\bm{\tau}_\mathrm{RBD}(t),\bm{q}(t),\bm{\dot{q}}(t),\bm{\ddot{q}}(t),\bm{r}(t))$, and 35 hidden neurons,
	\item the second LSTM cell has 35 input neurons and 7 hidden neurons, corresponding to the individual joint torques.
\end{itemize}
This network is abbreviated as \textbf{LSTM-2}. 

Based on this same topology, we implement the architectures \textbf{LSTM-FCL 100}  and \textbf{LSTM-FCL 500}, where a fully connected layer with 35 input and 7 output neurons with linear activation is used instead of the second cell to improve the regression performance by compensating for the squashing sigmoid output function of LSTM cells; the sequence lengths used for training are either $T=100$ or $T=500$.
\subsubsection{\textbf{Transformer}}\label{Transformer}
Originally proposed in \cite{vaswani}, the Transformer network is investigated as a non-recurrent encoder-decoder topology, having achieved favorable results in sequence-to-sequence modeling applications. The input is concatenated by a sinusoidal encoding and fed through a series of encoding layers, each containing a multi-head scaled dot-product attention, normalization, and fully connected sub-layer. The encoded features are then fed through decoder layers and are subjected to a joint attention layer accepting both the encoding and the previous predictions to generate the next output. The \emph{vanilla} implementation of the architecture as presented in the seminal work is adopted and configured for input sequence and output dimensions as defined for the LSTM, yielding the network function $f_\mathrm{TF}(\bm{x}({t_1}),\bm{x}({t_2}),\ldots,\bm{x}({T})) = \bm{\tau}_\mathrm{NN}$.
\subsubsection{\textbf{Input/Output Normalization}}\label{ioNorm}
Considering the largely varying numerical scales of the input features, all architectures are equipped with integral input and output normalization layers. Using the training data (see \ref{trainingMotion}), the mean $\overline{x_i}$ and standard deviations $\sigma(x_j)$ are computed over all observations of the input features $x_i \in \bm{x}$ in (\ref{inputSpace}) respectively to normalize each input $x_j$ according to
\begin{align}
	x_i^\mathrm{norm} = \frac{x_i - \overline{x}_i}{\sigma(x_i)}\;,
\end{align}
before being propagated forward through the network. The inverse operation is performed at the output where the prediction torques are subject to
$\tau ^\mathrm{pred}_j= \sigma(\tau_j)\tau^\mathrm{norm}_j+\overline{\tau}_j$.
\subsection{Algorithmic Training Motion Generation with LIMOPA}\label{trainingMotion}
Dynamic modeling is often performed for motion tasks within a fairly constrained effective workspace large-scale movement patterns \cite{vijayakumar2000locally,polyodros}, e.g., palletization or pick-and-place tasks, characterized by mainly joint-wise anisotropic motion. To our best knowledge, the dynamic modeling of collaborative robots over a wide kinematic during LIMO and fine-scale movements was not attempted before. To provide a data set for the development and evaluation of IDMs for this motion type, we propose a novel algorithmic method for the parametric generation of randomized LIMO-paths (see Algorithm \ref{alg:LIMOPA}). The aims are 1) to provoke hysteretic effects at low velocities and directional reversals within narrow positional windows, i.e., backlash and non-linear friction, and 2) to maximize the kinematic range over which these are observed.
\begin{algorithm}[b]
	\captionsetup{font={small}}
	\caption{Locally Isotropic Motion Path Algorithm (LIMOPA)}\label{alg:LIMOPA}
	\begin{algorithmic}[]
		\renewcommand{\algorithmicrequire}{\textbf{Input:}}
		\renewcommand{\algorithmicensure}{\textbf{Output:}}
		\scriptsize

		\While{numberOfScaffoldConfigs < goalNumberOfScaffoldConfigs}
		\State randomConfig = \textit{getRandomConfig}(jointIntervals)
		\State endEffectorPosition = \textit{forwardKinematics}(randomConfig)
		\State feasible = \textit{checkFeasibility}(randomConfig)
		\If{endEffectorPosition $\bm{\in}$ cartesianSampleRange \textbf{and} feasible == True}
		\State scaffoldConfigs[end] = randomConfig
		\EndIf
		\EndWhile
		\State sortedScaffoldConfigs[1] = scaffoldConfigs[1]
		\For{i=1 \textbf{to} \textit{length}(sortedScaffoldConfigs)}
		\State queriedConfig = sortedScaffoldConfigs[i]
		\State \textbf{remove} queriedConfig \textbf{from} scaffoldConfigs
		\State neighborConfig[i+1] = \textit{kdNearestNeighbor}(queriedConfig, scaffoldConfigs)
		\State \textbf{remove} neighborConfig \textbf{from} scaffoldConfigs
		\State sortedScaffoldConfig[i+1] = neighborConfig
		\State configPathPoints[end+1] = neighborConfig
		\State sinusoidals = \textit{generateRandomSinusoidals}()
		\State \textit{append}(configPathPoints, sinusoidals)
		\EndFor\\
		\Return configPathPoints
	\end{algorithmic}
\end{algorithm}
LIMOPA partitions the motion in configuration space into linear \emph{reaching}-phases where target configurations are approached, and subsequent sinusoidal \emph{exploring}-phases where the robot performs locally isotropic reconfigurations around the prior target configuration. 
\begin{figure}[t]
	\centering
	\includegraphics[width=1\columnwidth]{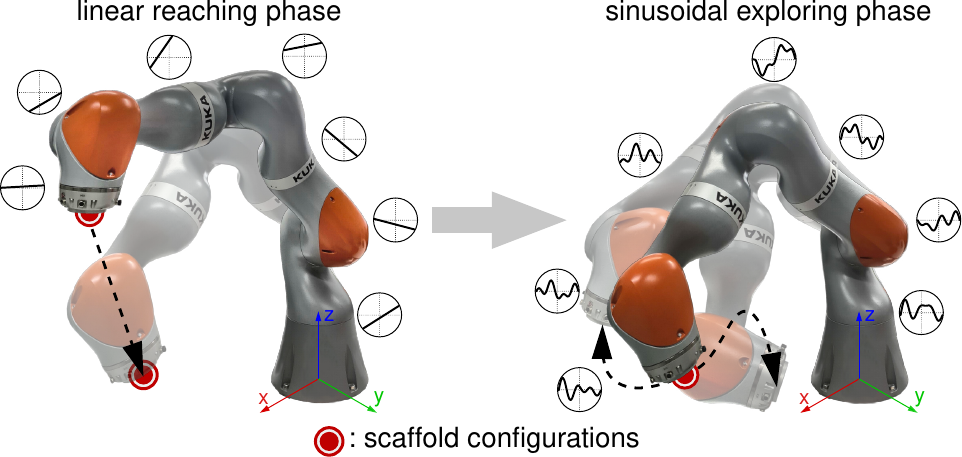}
	\caption{Path geometry generated by LIMOPA.
		(\textit{Left}) After interpolation between two scaffold configurations (see red markers) via random, joint-wise linear motion profiles, (\textit{Right}) low-amplitude randomized sinusoidal joint trajectories are appended around the current scaffold configuration.}
	\label{fig}
\end{figure}
This is realized by the pre-calculation of a geometric path  of length $P$ composed as a joint configuration sequence $\bm{C}=\{\bm{q}_1,\ldots,\bm{q}_P\}$ with $\bm{q}_p \in \mathbb{R}^{n=7}$ as the $p$-th waypoint of $\bm{C}$. The implementation is based on a randomly generated list of kinematically feasible and collision-free \emph{scaffold}-configurations $\bm{S} =\{\bm{q}^{\mathrm{scaf}}_1,\ldots,\bm{q}^{\mathrm{scaf}}_L\} $ of length $L$, that are identified in simulation. The scaffold configuration $ \bm{q}^{\mathrm{scaf}}_l $ represents the goal configuration to be reached in the $l$-th reaching-phase via joint-wise linear motion. This is followed by the $ l $-th exploration-phase in the form of a time-unfolded joint-wise random sinusoidal path $ _j\bm{e}_l \in \mathbb{R}^T$ of length $T$ parameterized by a discrete Fourier series
\begin{align}
	&_je_l({t}) = \sum^3_{k=1} A_k^{\mathrm{rand}} \mathrm{sin}\left( 2\pi f_k^{\mathrm{rand}} t + \varphi_k^{\mathrm{rand}}\right)\;,\\
	&_j\bm{e}_l = \{_je_l(1),\ldots,{_je_l(T^{\mathrm{rand}})}\}\;.
\end{align}
$ A_k^{\mathrm{rand}}\in [0.5,3] $, $ f_k^{\mathrm{rand}} \in [-4,4] $ and $ \varphi_k^{\mathrm{rand}} \in [0,2\pi] $ are drawn randomly from the respective intervals for each joint $j$ and each $ k $-th summand of the Fourier series. The path length of the $ l $-th exploration phase $ \bm{e}_l =\left( {{_1\bm{e}_l},\ldots,{_{n=7}\bm{e}_l}}\right) \in \mathbb{R}^{(n=7)\times T^{\mathrm{rand}}}$ is randomly determined by $T^{\mathrm{rand}} \in [2000,2500]$. This results in a configuration path
\begin{align}
	\bm{C} = \{\underbrace{\bm{q}_1^{\mathrm{scaf}}}_{{7\times 1 }},\underbrace{\bm{e}_1}_{{7\times T^{\mathrm{rand}}}},\underbrace{\bm{q}_2^{\mathrm{scaf}}}_{{7\times 1 }},\underbrace{\bm{e}_2}_{{7\times T^{\mathrm{rand}}}},\ldots,\underbrace{\bm{q}_L^{\mathrm{scaf}}}_{{7\times 1 }},\underbrace{\bm{e}_L}_{{7\times T^{\mathrm{rand}}}}\}\;.
\end{align}
Joint-wise independent, random sinusoidal motions are chosen to excite a wide range of dynamic state combinations, also provoking hysteretic behavior via sinusoidal oscillations; the parameter intervals are defined to preserve nonlinearity by preventing saturation of potential hysteresis loops.
Although neural network-based approaches rely on the availability of a rich and representative training data set, the high dimensionality of robot dynamics, encompassing at least $\mathrm{dim}(\bm{q},\bm{\dot{q}},\bm{\ddot{q}},\bm{\tau}) = 4\times n$ state space dimensions for $n$ joints, makes an exhaustive dynamic state space sampling intractable. To balance sampling efficiency and representative quality, we define an effective workspace $W_\mathrm{eff}$ as the kinematically reachable quadrant of a sphere radius 0.8 m in front of the robot, bounded by the y-axis and the positive x- and z-axes of the base coordinate system to be sampled as a subspace of the reachable workspace $W_\mathrm{tot}\supset W_\mathrm{eff}$ within which local manipulator dexterity is preserved \cite{callar,ivo}.

\begin{figure}[t]
	\centerline{\includegraphics[width=1\columnwidth]{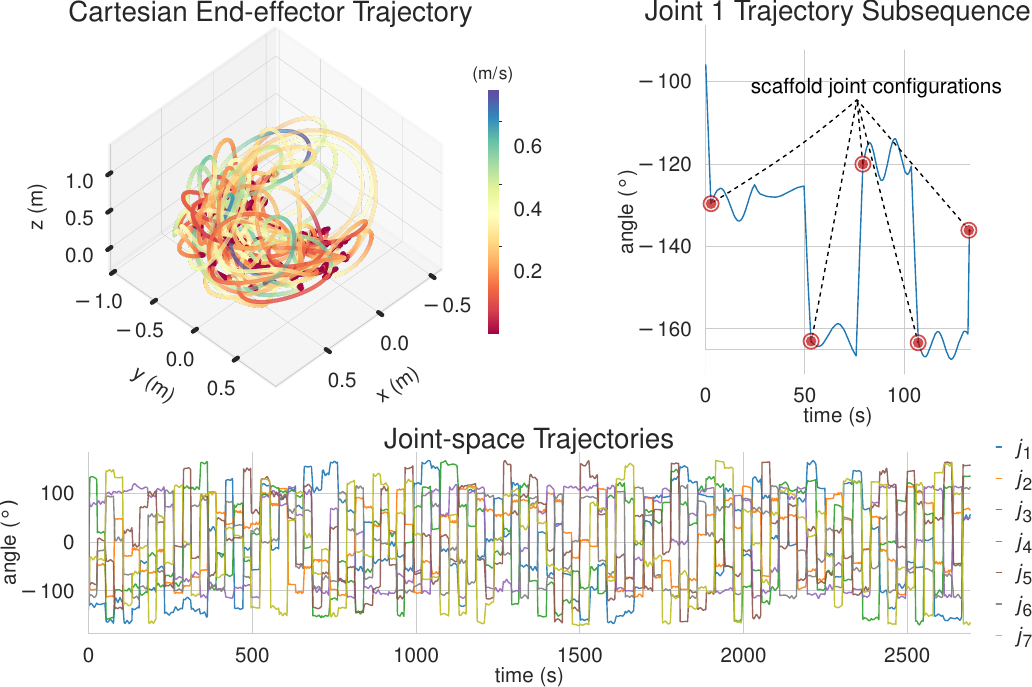}}
	\caption{(\emph{Top left}) Trajectory traversed by the end-effector while executing the training motion path generated by LIMOPA (\ref{trainingMotion}). Note the red segments where joint-wise locally isotropic sinusoidal motions are executed. (\emph{Top right}) A section of the sinusoidal trajectory executed on joint 1 shows locally isotropic motion in configuration space. (\emph{Bottom}) The total training trajectory in configuration space.}
	\label{jointSpaceTrajectory}	 	
\end{figure}
\begin{figure}[h]
	\centerline{\includegraphics[width=1\columnwidth]{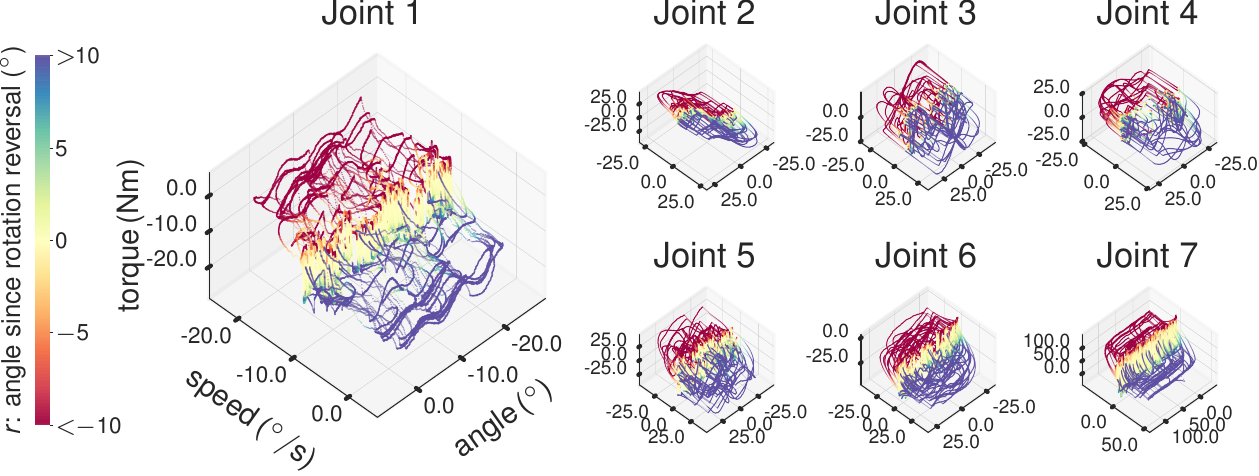}}
	\caption{Visualization of the dynamic space of each joint spanning angle, speed, torque and rotation since direction reversal contained in the training data.
	}
	\label{jointSpace} 	

\end{figure}
Given the robot joint limits, the joint spaces are uniformly sampled to obtain candidate configurations subject to a feasibility check for collisions and compliance with the sample space definition in simulation. This is repeated until, in our case, 150 so-called \textit{scaffold} configurations are identified. To reduce the path execution time on the real robot, the scaffold configurations are sorted by distance in configuration space via a nearest-neighbor search based on a kd-tree representation of the configuration space.
As described, the final geometric path is lastly constructed by inserting an exploration sequence after each scaffold configuration. With an average of 2250 waypoints for each exploration phase, this results in approximately, 337500 waypoints for the total path and an execution duration of roughly 1 hour.
The waypoints on the geometric path are sequentially executed by the proprietary controller of the robot manufacturer such that the waypoint transitions are executed as linear configuration space interpolations with soft real-time constraints. At the beginning of each reaching phase, the joint-wise relative velocity and acceleration are randomly set between 60\% and 90\% with respect to the individual maximum joint velocities. During exploring phases, the relative velocities are randomly set between 10\% and 30\%.

Through this motion generation strategy, thorough coverage of the Cartesian and configuration workspace is achieved. Concurrently, the coverage of a wide range of dynamic state combinations in the acquired data is ensured by including multiple permutations of simultaneously occurring kinematic states, motion directions, and torques, yielding a comprehensive yet efficient sampling of the dynamic state space (Fig. \ref{jointSpace} and \ref{jointSpaceTrajectory}). We provide our LIMOPA data set under \cite{repo}.

\subsection{Training Scheme}
The neural networks are trained on the data set shown in Fig. \ref{jointSpaceTrajectory} containing 804205 sequential joint positions and torques, where the velocities and accelerations are obtained by numeric differentiation. 
 
We employ a random train-test-split of 80\% to 20 \% for the data set. With respect to the time-series networks, the data set is partitioned into sub-sequences, such that the shuffling affects only the order of the sub-sequences but not that of the samples within one such sub-sequence.
The training of all networks is conducted for 10 runs with Xavier- and zero-initialization for 100 epochs using the AdamW optimizer \cite{loshchilov2019decoupled} with an empirically determined initial learning rate $\epsilon = 0.001$ that is reduced when the training loss does not decrease for consecutive epochs. To prevent over-fitting, the batch size is set to 50.
The mean squared error
\begin{align}
	L_{\mathrm{ MSE}}(\bm{\tau}_\mathrm{HYB},\widetilde{\bm{\tau}}_\mathrm{meas}) = \frac{1}{m}\sum_{i}^{m}\lvert\lvert\left(\bm{\tau}_\mathrm{HYB}- \widetilde{\bm{\tau}}_\mathrm{meas}\right)_i^2\rvert\rvert\;,
	\label{trainingLoss}
\end{align}
between $ \bm{\tau}_\mathrm{HYB}$ and the low-pass filtered torque measurements $\widetilde{\bm{\tau}}_\mathrm{meas}$, i.e., to reduce torque ripples due to vibrations induced by nonlinear actuator transmission \cite{8206543}, is used as the loss metric. The prediction modules are trained within the hybrid model structure (see (\ref{hybridArchitecture})) either {1}) with the full input feature space defined in (\ref{inputSpace}), {2}) as standalone networks without the input feature inclusion of $\bm{\tau}_\mathrm{RBD}$ and the output addition of the upstream RBD model estimates, {3}) without the rotational history encoding input $ \bm{r} $, or in case of the time-series networks 4) with different input sequence lengths $ T=\{100,500\} $. As Transformers are intended for long sequence lengths exceeding the hypothetical capability of recurrent topologies, the Transformer is only trained once on sequence lengths of $ T=500 $ and with the full feature space.
\section{Evaluation of the Model Architectures}\label{eval}

To explore optimal IDM architecture and configuration for LIMO dynamics, we focus on the implications of various components of the base architecture presented in Section \ref{hybridModelLeraningBaseArchitecture}.

The performance of the proposed model architectures and their variations are evaluated comparatively on the training and test data sets acquired during the execution of LIMO according to the trajectory generated by LIMOPA (Sec. \ref{trainingMotion}). 
The models are further validated on a separate data set acquired during the execution of a sequence of large-amplitude sinusoidal excitation trajectories (akin to the trajectory used for parametric regression) over the full workspace with decreasing base frequency to obtain dynamic and hysteretic states not contained in the original training data set, e.g. large accelerations, speeds, and rotations in one direction. 

\begin{table}[b]
	\caption{MSE between model predictions and filtered measured torques. The results are sorted joint-wise by color.}\label{numericResults}
	\label{mseJoints}
	\begin{adjustbox}{width=\columnwidth}
		\centering
		\begin{NiceTabular}{r l S[table-format=1.4] S[table-format=3.4] S[table-format=3.4]    S[table-format=2.4]   S[table-format=1.4] S[table-format=1.4] S[table-format=1.4] S[table-format=2.4]}
			
			\\\specialrule{.4pt}{0pt}{1pt}
			\multirow{15}{*}{\begin{turn}{90}
					\textbf{Training MSE (Nm)}
			\end{turn}} &\multicolumn{1}{l}{\textbf{Architecture}}& {\textbf{Joint 1}} & {\textbf{Joint 2}} &{\textbf{Joint 3}} &{\textbf{Joint 4}} &{\textbf{Joint 5}} &{\textbf{Joint 6}} &{\textbf{Joint 7}} & {\textbf{Joint Avg.}}
			\\\specialrule{.4pt}{0pt}{1pt}
			&{RBD } &    \cellcolor{14} 1.9499  & \cellcolor{12}  25.2967        &   \cellcolor{12}  0.8434      &   \cellcolor{12}  2.7134    &    \cellcolor{13}  0.3373   &   \cellcolor{9}  0.0963    &       \cellcolor{9}0.0442    &  \cellcolor{12}4.4687    \\
			
			&{MLP-7} &      \cellcolor{13}  1.0049 &     \cellcolor{14} 202.5057&      \cellcolor{14} 179.1369&   \cellcolor{14}    77.0004&  \cellcolor{14}     0.4084&  \cellcolor{14}     0.5081&  \cellcolor{13}     0.1949  &  \cellcolor{14} 65.8228\\
			
			&{MLP-7 with $ \bm{\tau}_\mathrm{RBD} $ } &      \cellcolor{10}0.4154 &     \cellcolor{11}  0.9537   &     \cellcolor{11}  0.3475   &      \cellcolor{10} 0.3101 &       \cellcolor{9}0.0468 &   \cellcolor{10}    0.1345 &     \cellcolor{10}  0.0701  &    \cellcolor{11} 0.3256    \\
			
			&{MLP-7 with $\bm{\tau}_\mathrm{RBD},\bm{r}$ } &  \cellcolor{9}0.2938 &      \cellcolor{10} 0.9160   &     \cellcolor{10}  0.3455   &    \cellcolor{11}   0.3115 &       \cellcolor{6}0.0251 &    \cellcolor{8}   0.0523 &     \cellcolor{7}  0.0387   &  \cellcolor{10}   0.2833\\
			
			&{LSTM-2 100} &\cellcolor{12} 0.6039 &   \cellcolor{13}   82.8548     &   \cellcolor{13} 12.9735       &  \cellcolor{13}  12.1421     &    \cellcolor{12} 0.2604    &    \cellcolor{13} 0.4004    &       \cellcolor{14} 0.2052    &  \cellcolor{13}  15.6344    \\
			
			&{LSTM-2 100 with $ \bm{\tau}_\mathrm{RBD} $ } &      \cellcolor{5} 0.0593   &    \cellcolor{4} 0.0952      &      \cellcolor{4}0.0207     &     \cellcolor{4} 0.0208  &     \cellcolor{5} 0.0214   &   \cellcolor{4} 0.0252     &    \cellcolor{5}  0.0234      &      \cellcolor{4}0.0380 \\
			
			&{LSTM-2 100 with  $\bm{\tau}_\mathrm{RBD},\bm{r}$ } &     \cellcolor{4} 0.0518    &  \cellcolor{2}   0.0517      &       \cellcolor{1}0.0163    &   \cellcolor{1}0.0171     &    \cellcolor{1} 0.0129    &    \cellcolor{1} 0.0124    &      \cellcolor{1}0.0111                &   \cellcolor{2}0.0248
			\\
			&{LSTM-FCL 100 } &     \cellcolor{8} 0.1138    &      \cellcolor{8} 0.1530    &    \cellcolor{9} 0.0694      &    \cellcolor{8} 0.0662    &  \cellcolor{11} 0.0735      &   \cellcolor{12}0.2152      &    \cellcolor{12} 0.1912     &  \cellcolor{9}  0.1260   \\
			
			&{LSTM-FCL 100 with $\bm{r}$ } &     \cellcolor{7} 0.0770    &      \cellcolor{9} 0.2207    &    \cellcolor{7} 0.0586      &    \cellcolor{7} 0.0567    &  \cellcolor{7} 0.0270      &   \cellcolor{7}0.0495      &    \cellcolor{8} 0.0413     &  \cellcolor{7}  0.0758   \\
			
			&{LSTM-FCL 100 with $\bm{\tau}_\mathrm{RBD},\bm{r}$} &      \cellcolor{3} 0.0385   &      \cellcolor{3}0.0683     &        \cellcolor{2}0.0188   &    \cellcolor{2}0.0198     &    \cellcolor{3} 0.0161    &   \cellcolor{3}0.0155      & \cellcolor{2}  0.0115       &   \cellcolor{3} 0.0270  \\
			
			&{LSTM-FCL 500} &      \cellcolor{6} 0.0713   &    \cellcolor{7}   0.1468    & \cellcolor{8} 0.0648      & \cellcolor{9}0.0698        &    \cellcolor{10} 0.0690    &  \cellcolor{11}  0.2011     &   \cellcolor{11} 0.1816       &    \cellcolor{8} 0.1150    \\
			
			&{LSTM-FCL 500 with $ \bm{\tau}_\mathrm{RBD} $} &  \cellcolor{2}  0.0317      &        \cellcolor{6}0.1177   &   \cellcolor{5} 0.0237       &  \cellcolor{5} 0.0249      &    \cellcolor{8} 0.0284    &     \cellcolor{5}  0.0265  & \cellcolor{6}    0.0300     &    \cellcolor{5} 0.0404    \\
			
			&{LSTM-FCL 500 with  $\bm{\tau}_\mathrm{RBD},\bm{r}$} &     \cellcolor{1}  0.0292   &     \cellcolor{1} 0.0507     &       \cellcolor{3}0.0196    &    \cellcolor{3}0.0199     &    \cellcolor{2} 0.0159    &   \cellcolor{2} 0.0133     &    \cellcolor{3} 0.0119      &   \cellcolor{1}  \textbf{ 0.0230}   \\
			
			&{Transformer 500 with  $\bm{\tau}_\mathrm{RBD},\bm{r}$} &        \cellcolor{11}0.4293 &      \cellcolor{5}0.1134   &     \cellcolor{6}  0.0459     &     \cellcolor{6} 0.0455   &    \cellcolor{4} 0.0190    &  \cellcolor{6} 0.0306      &    \cellcolor{4} 0.0216     &   \cellcolor{6} 0.0461  \\\specialrule{.4pt}{1pt}{1pt}

			\multirow{14}{*}{\begin{turn}{90}
					\textbf{Test MSE (Nm)}
			\end{turn}}&{RBD }  & \cellcolor{14}2.0882    &     \cellcolor{12} 23.1839     &     \cellcolor{8} 0.0991     &  \cellcolor{12} 2.1749      &  \cellcolor{14} 0.4097      &   \cellcolor{9} 0.0985     &   \cellcolor{8} 0.0471      & \cellcolor{12}4.0145\\
			
			&{MLP-7} &\cellcolor{13}0.9836&    \cellcolor{14}   207.1595   &     \cellcolor{14}  178.9357   &     \cellcolor{14}  75.4605&       \cellcolor{13}0.4061 &       \cellcolor{14}0.5070& \cellcolor{13}      0.196& \cellcolor{14} 66.2355  \\
			
			&{MLP-7 with $ \bm{\tau}_\mathrm{RBD} $ } & \cellcolor{11}0.4081 &     \cellcolor{11}  0.9160   &    \cellcolor{12}   0.3758&     \cellcolor{11}  0.3148&      \cellcolor{9} 0.0463&     \cellcolor{10}  0.1360 &       \cellcolor{10}0.0710 & \cellcolor{11}0.3240          \\
			
			&{MLP-7 with $\bm{\tau}_\mathrm{RBD},\bm{r}$ }&      \cellcolor{10} 0.2747 &     \cellcolor{10}  0.8845   &      \cellcolor{11} 0.3309   &     \cellcolor{10}  0.2916 &      \cellcolor{6} 0.0252 &   \cellcolor{1}    0.0054 &      \cellcolor{7} 0.0385& \cellcolor{10}0.2644\\
			
			&{LSTM-2 100} & \cellcolor{12}0.6095     &   \cellcolor{13} 80.8355       &     \cellcolor{13}12.3105      &  \cellcolor{13}11.5995       &    \cellcolor{12} 0.2737    &   \cellcolor{13} 0.4068   &   \cellcolor{14} 0.2058   &   \cellcolor{13} 15.1774   \\
			
			&{LSTM-2 100 with $ \bm{\tau}_\mathrm{RBD} $ }& \cellcolor{5}0.0730   &      \cellcolor{4}0.1297     &    \cellcolor{5} 0.0321      &  \cellcolor{5} 0.0338      &    \cellcolor{5}0.0239     &    \cellcolor{6} 0.0291    &    \cellcolor{5} 0.0262 &   \cellcolor{5} 0.0497     \\
			
			&{LSTM-2 100 with  $\bm{\tau}_\mathrm{RBD},\bm{r}$ } &     \cellcolor{4} 0.0663   &       \cellcolor{2}0.0849    &     \cellcolor{3}0.0274      &  \cellcolor{2}0.0266       &     \cellcolor{1}0.0151    &  \cellcolor{2}0.0138       &   \cellcolor{1} 0.0121 &  \cellcolor{2} 0.0352     \\
			
			&{LSTM-FCL 100 } &     \cellcolor{8} 0.1254    &      \cellcolor{7} 0.2114    &    \cellcolor{7} 0.0886      &    \cellcolor{7} 0.0922    &  \cellcolor{11} 0.0807      &   \cellcolor{12}0.2282      &    \cellcolor{12} 0.1924     &  \cellcolor{9}  0.1455   \\
			
			&{LSTM-FCL 100 with $\bm{r}$ }& \cellcolor{7}0.1224   &     \cellcolor{8} 0.2955     &    \cellcolor{9}  0.1092     &     \cellcolor{6}0.0777    &   \cellcolor{8} 0.0315     &  \cellcolor{8} 0.0595      &   \cellcolor{9}  0.0487&  \cellcolor{6}0.1064    \\
			
			&{LSTM-FCL 100 with $\bm{\tau}_\mathrm{RBD},\bm{r}$}& \cellcolor{3}0.0526    &     \cellcolor{3} 0.0991     &    \cellcolor{2}0.0258       &   \cellcolor{1} 0.0265     &    \cellcolor{3}0.0171     &   \cellcolor{4} 0.0181     &   \cellcolor{2}  0.0124 &     \cellcolor{3}0.0360    \\
			
			&{LSTM-FCL 500} & \cellcolor{6}0.0754   & \cellcolor{6}  0.1842        &   \cellcolor{6} 0.0798       &  \cellcolor{8} 0.0954      &    \cellcolor{10}0.0740     &    \cellcolor{11}  0.2200   &   \cellcolor{11} 0.1825 &\cellcolor{8} 0.1302    \\
			
			&{LSTM-FCL 500 with $ \bm{\tau}_\mathrm{RBD} $} &\cellcolor{2}0.0410   &      \cellcolor{5}0.1380     &   \cellcolor{4} 0.0309       &    \cellcolor{4} 0.0319    &    \cellcolor{7}  0.0293   &  \cellcolor{5} 0.0273      &     \cellcolor{6}0.0319&  \cellcolor{4}0.0472   \\
			
			&{LSTM-FCL 500 with  $\bm{\tau}_\mathrm{RBD},\bm{r}$} & \cellcolor{1}0.0400   &   \cellcolor{1}   0.0628    &     \cellcolor{1} 0.0253     &   \cellcolor{3} 0.0271     &   \cellcolor{2} 0.0168     &  \cellcolor{3}0.0141       &    \cellcolor{3}   0.0124 &\cellcolor{1} \textbf{ 0.0284}  \\
			
			&{Transformer 500 with  $\bm{\tau}_\mathrm{RBD},\bm{r}$} &\cellcolor{9} 0.1432     &    \cellcolor{9} 0.3179      &   \cellcolor{10}0.1222        & \cellcolor{9}0.1284        &     \cellcolor{4}  0.0236  &  \cellcolor{7} 0.0390      &   \cellcolor{4}0.0244 &  \cellcolor{7}  0.1142      
			\\\specialrule{.4pt}{1pt}{1pt}

			\multirow{14}{*}{\begin{turn}{90}
					\textbf{Validation MSE (Nm)}
			\end{turn}}&{RBD }  & \cellcolor{1}0.3487    &     \cellcolor{10} 31.0620     &     \cellcolor{1} 2.6918     &  \cellcolor{4} 4.7038      &  \cellcolor{3} 0.0639      &   \cellcolor{1} 0.0833     &   \cellcolor{1} 0.0177      & \cellcolor{10}5.5673\\
			
			&{MLP-7} &\cellcolor{13}2.4092&    \cellcolor{13}   234.5064   &     \cellcolor{14}  100.1022   &     \cellcolor{14}  120.5577&       \cellcolor{13}0.8988 &       \cellcolor{11}0.6676& \cellcolor{8}      0.0627& \cellcolor{14} 66.6006  \\
			
			&{MLP-7 with $ \bm{\tau}_\mathrm{RBD} $ } & \cellcolor{4}0.4618 &     \cellcolor{2}  10.3878   &    \cellcolor{6}   4.2824&     \cellcolor{3}  4.0479&      \cellcolor{4} 0.0644&     \cellcolor{3}  0.1134 &       \cellcolor{3}0.0387 & \cellcolor{2}2.7709          \\
			
			&{MLP-7 with $\bm{\tau}_\mathrm{RBD},\bm{r}$ }&      \cellcolor{7} 0.4289 &     \cellcolor{1}  7.4943   &      \cellcolor{3} 3.6548   &     \cellcolor{2}  3.8454 &      \cellcolor{1} 0.0386 &   \cellcolor{2}    0.1056 &      \cellcolor{2} 0.0207& \cellcolor{1}\textbf{ 2.2269}\\
			
			&{LSTM-2 100} & \cellcolor{7}1.5878     &   \cellcolor{14} 304.6905       &     \cellcolor{13}57.9000      &  \cellcolor{13}88.5953       &    \cellcolor{11} 0.5772    &   \cellcolor{12} 0.8465   &   \cellcolor{11} 0.0881   &   \cellcolor{13} 64.8979   \\
			
			&{LSTM-2 100 with $ \bm{\tau}_\mathrm{RBD} $ }& \cellcolor{5}0.5207   &      \cellcolor{7}16.3673     &    \cellcolor{4} 4.0213      &  \cellcolor{7} 5.8706      &    \cellcolor{6}0.0802     &    \cellcolor{9} 0.3689    &    \cellcolor{5} 0.0432 &   \cellcolor{4} 3.8960     \\
			
			&{LSTM-2 100 with  $\bm{\tau}_\mathrm{RBD},\bm{r}$ } &     \cellcolor{3} 0.4342   &       \cellcolor{5}15.1118    &     \cellcolor{5}4.2252      &  \cellcolor{5}5.2702       &     \cellcolor{5}0.0743    &  \cellcolor{7}0.2406       &   \cellcolor{9} 0.0627 &  \cellcolor{3} 3.6313     \\
			
			&{LSTM-FCL 100 } &     \cellcolor{11} 1.9988    &      \cellcolor{11} 92.3633    &    \cellcolor{11} 43.3894      &    \cellcolor{12} 63.3062    &  \cellcolor{12} 0.8600      &   \cellcolor{13}1.2037      &    \cellcolor{12} 0.1299     &  \cellcolor{11}  29.1216   \\
			
			&{LSTM-FCL 100 with $\bm{r}$ }& \cellcolor{14}2.4335   &     \cellcolor{6} 15.4618     &    \cellcolor{7}  4.6502     &     \cellcolor{9}7.7476    &   \cellcolor{9} 0.2056     &  \cellcolor{10} 0.4692      &   \cellcolor{14}  0.2863&  \cellcolor{8}4.4649    \\
			
			&{LSTM-FCL 100 with $\bm{\tau}_\mathrm{RBD},\bm{r}$}& \cellcolor{9}1.6668    &     \cellcolor{3} 12.8295     &    \cellcolor{8}5.6193       &   \cellcolor{10} 7.7479     &    \cellcolor{10}0.2303     &   \cellcolor{5} 0.1847     &   \cellcolor{7}  0.0616 &     \cellcolor{6}4.0486    \\
			
			&{LSTM-FCL 500} & \cellcolor{12}2.1022   & \cellcolor{12}  177.2180        &   \cellcolor{12} 52.5657       &  \cellcolor{11} 48.5196      &    \cellcolor{14}0.9719     &    \cellcolor{14}  1.9562   &   \cellcolor{13} 0.1507 &\cellcolor{12} 40.4978    \\
			
			&{LSTM-FCL 500 with $ \bm{\tau}_\mathrm{RBD} $} &\cellcolor{8}1.6170   &      \cellcolor{8}18.0963     &   \cellcolor{10} 6.5467       &    \cellcolor{6} 5.7261    &    \cellcolor{8}  0.1450   &  \cellcolor{6} 0.2118      &     \cellcolor{10}0.0810&  \cellcolor{9}4.6320   \\
			
			&{LSTM-FCL 500 with  $\bm{\tau}_\mathrm{RBD},\bm{r}$} & \cellcolor{10}1.7519   &   \cellcolor{4}   13.6668    &     \cellcolor{9} 6.2471     &   \cellcolor{8} 6.4873     &   \cellcolor{7} 0.1268     &  \cellcolor{8}0.2433       &    \cellcolor{4}   0.0425 &\cellcolor{7} 4.0808  \\
			
			&{Transformer 500 with  $\bm{\tau}_\mathrm{RBD},\bm{r}$} &\cellcolor{6} 0.8202     &    \cellcolor{9} 20.3129      &   \cellcolor{2}3.3142        & \cellcolor{1}3.6276        &     \cellcolor{2}  0.0561  &  \cellcolor{4} 0.1471      &   \cellcolor{6}0.0448 &  \cellcolor{5}  4.0461\\\specialrule{.4pt}{1pt}{0pt}
		\end{NiceTabular}
	\end{adjustbox}\vspace{-0cm}
\end{table}
To this end, an inexhaustive grid search is performed by ablating parts of the input feature space and connections between the model components as per (\ref{inputSpace}) and Fig. \ref{fig:NetArch}, and variation of the neural network base architecture.
The tested modules are named after their base architecture (see Sec. \ref{baseArchitecture}) and the specifiers "with $\bm{r}$", indicating usage of the rotational history encoding $\bm{r}$, and "with $\bm\tau_{\mathrm{RBD}}$", denoting the model being configured as a hybrid architecture where $\bm\tau_{\mathrm{RBD}}$ is used both as a neural network input and for the final output addition (Fig. \ref{hybridArchitecture}). Thus, the architectures are rendered equivalent to either the sole baseline RBD model, neural network component, or the full hybrid architecture.

\begin{figure*}[]
	\adjustbox{width=\textwidth}{
		\includegraphics[]{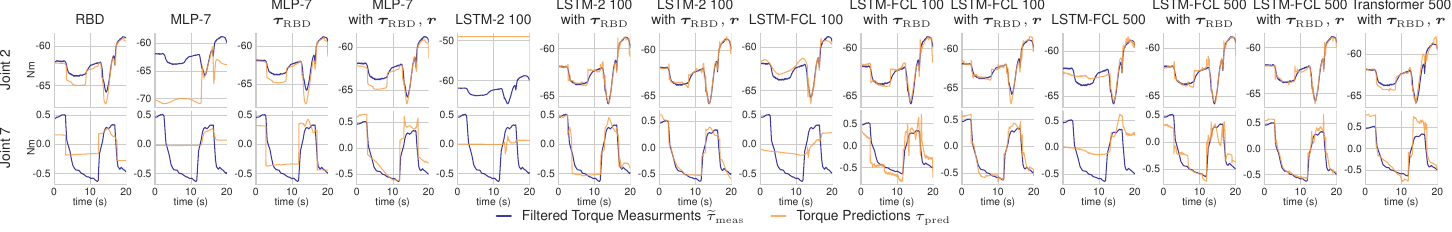}
	}\centering
	\caption{Torque predictions of the evaluated model architectures during a LIMOPA-based trajectory for joints 2 and 7.}
	\label{inferences}
\end{figure*}
\begin{figure}[b]
	\includegraphics[width=\columnwidth]{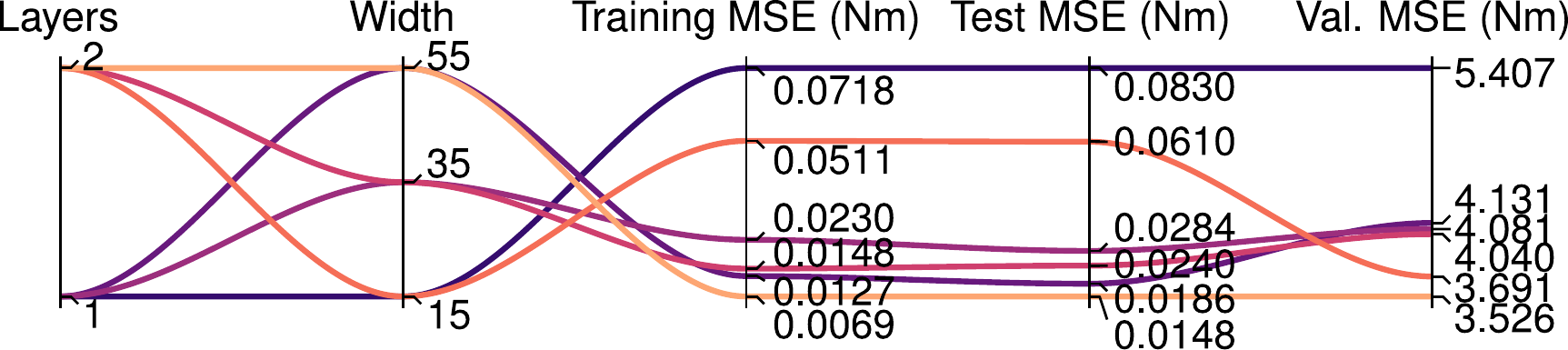}
	\centering
	\caption{MSEs of variations of LSTM-FCL-500 with $\bm\tau_{\mathrm{RBD}},\bm{r}$ investigated within ablation experiments.}
	\label{parallelplot}
\end{figure}
On all data sets, models trained in hybrid configuration consistently achieved lower MSEs than their standalone counterparts, where hybridization of MLP-7 and LSTM-2 100 lead to MSE reductions by factors in the magnitude of $ 10^3$ (see Table 1).  This substantiates previous claims of a boosted absolute model performance through grey-box approaches, or at least better convergence on limited training data. In view of related hybrid modeling approaches carried out on dynamically similar robots as in \cite{hitzler}, the problem of underfitting on highly dynamic robotic motion data seems common, necessitating the inclusion of model priors like an RBD model.

Interestingly, however, LSTM-FCL and Transformer architectures, converged to low MSEs on the LIMO data set even in non-hybrid configuration, whereby in hybrid configuration, LSTM-FCL achieved the lowest MSE. Specifically, the highest accuracy for the 500 variant, indicates that longer sequences may still contain additional exploitable information. The high accuracy of LSTM-FCL, may originate from its recurrent architecture and regression ability via the distal fully connected layer. The Transformer also evaluated on sequences of length $ T=500 $ yielded no further improvement, which may indicate an advantage of recurrence over attention mechanisms. 

A grid search for the optimization of topological hyperparameters for the best-performing model LSTM-FCL 500 with $\bm\tau_{\mathrm{RBD}},\bm{r}$ revealed both the addition of a second LSTM cell and the increase of the cell width to independently provide accuracy gains (Fig. \ref{parallelplot}). An exception was observed on the validation data where two layers with 15 neurons performed closely to two layers and 55 neurons and outperformed the other variants. As this is not seen in the LIMO data sets,

we suspect a lower dimensionality due to a width of 15 over two layers to cause a structural bias-variance balance allowing to rather generalize to non-LIMO dynamics contained in the validation data by restricting encoding of LIMO dynamics.

In terms of further utilization of sequence history information, each model trained with the rotational history encoding $\bm{r}$ outperformed their counterparts, demonstrating the benefit of sequence history information for non-recurrent and recurrent topologies alike.
Respecting the performance on the non-LIMO validation data set, we report a reduced, albeit still significant, benefit of $\bm{r}$. This can be regarded as evidence for our assumption, that sequence-history effects, i.e. hysteresis, play a more critical role during LIMO in contrast to conventional motion.
This is also supported by a) the accuracy gains attributable to the use of time-series architectures on the LIMO data set, where these are able to leverage more temporal information, and, conversely, b) relative accuracy losses on the non-LIMO validation data set. Likewise, the RBD performed similarly on all data sets, but significantly worse than the remaining hybrid models during LIMO. This again may be due to the proportionally higher influence of hysteretic effects being mainly provoked by LIMO dynamics outside the framework of RBD. 
Surprisingly, the hybrid MLP architectures performed best on the non-LIMO validation data in accordance with \cite{yilmaz,hitzler} where this model class was successfully used for IDM of non-LIMO. We suspect predominant exposure to LIMO data during training leading the specialized time-series architectures to adapt to this motion type. Although re-training on data containing other dynamic ranges may be necessary to improve model generalization, the aforementioned topological optimization (Fig. \ref{parallelplot}) already yielded significant accuracy gains on the validation data set.

The temporal courses of the model predictions in Fig. \ref{inferences} show the qualitative superiority obtained by model hybridism and time-series encoding capabilities by the congruence between predictions and measurements across all joints. In particular, this is apparent at inflection points in the torque profiles at directional reversals of joint 7, which is structurally isolated from gravitational torques, e.g. for the MLPs at $t\approx[2s,12s]$: Without $\bm{r}$ the prediction curve is of rectangular shape with sudden jumps. This is mitigated by the rotation history encoding used by MLP-7 with $r$, which is able to approximate the slope of the torque curve. This qualitative advantage is present in the remaining time-series architectures as well, where the temporal information further reduces overshooting predictions at motion reversals and close tracking of the true torques, even for joint 2 with the highest dynamic range. 

\section{Conclusion}
The inverse dynamics modeling problem is presented, emphasizing the complexity of modeling slow, locally isotropic, for which we postulated a relative increase in the contribution of a plethora of non-linear phenomena to the total dynamics originating from joint flexibility and low-velocity friction. Parametric model identification proved unsatisfactory in capturing the dynamics of motion sequences that provoke said effects. Likewise, conventional neural networks were shown to not yield satisfactory results in face of the data scarcity. Using time-series model learning based on LSTM and Transformer network topologies, we demonstrated that exploiting temporal information provides substantial model accuracy gains, that could be increased and transferred to non-recurrent architectures through the introduction of a rotational history encoding.
We provided a hybrid base architecture that leverages the combination of a physics-based model prior with the universal approximation ability of neural networks, with three main advantages over conventional black or white box approaches for dynamic modeling: Firstly, the physical explainability of model predictions is retained. Secondly, the needed amount of training data could be reduced. Lastly, significant increases in joint torque estimation accuracy during complex and conventional motion were achieved. As a next step, we intend to investigate the integration of our model architecture within a real-time force control scheme.
%
%
%

\ifCLASSOPTIONcaptionsoff
\newpage
\fi

	\bibliographystyle{IEEEtran}
	\bibliography{IEEEabrv,callarBIB}


\end{document}